\documentclass[journal]{IEEEtran}

\ifCLASSOPTIONcompsoc
  \usepackage[nocompress]{cite}
\else

\usepackage{cite}

\fi

\usepackage{pgfplotstable}
\usepackage{afterpage}
\usepackage{lscape}
\usepackage{enumitem}
\usepackage{multirow}

\usepackage{mathtools}
\newcommand{\matr}[1]{\mathbf{#1}}

\usepackage{etoolbox}
\robustify\bfseries
\newcommand{\sbf}{\bfseries}
\usepackage{siunitx}
\DeclareSIUnit\pixel{px}

\ifCLASSOPTIONcompsoc
 \usepackage[caption=false,font=normalsize,labelfont=sf,textfont=sf]{subfig}
\else
 \usepackage[caption=false,font=footnotesize]{subfig}
\fi

\usepackage[numbers]{natbib}

\usepackage{pgfplots}
\usepackage{tikz}
\usetikzlibrary{calc}

\usepackage{xcolor}
\usepackage{booktabs}
\usepackage{url}
\usepackage{color}
\usepackage{todonotes}
\usepackage{tabularx}
\usepackage{caption}

\usepackage{afterpage}
\usepackage{enumitem}

\usepackage{epstopdf}
\AppendGraphicsExtensions{.tif}
\AppendGraphicsExtensions{.jpg}

\definecolor{tumblue}{RGB}{0,101,189} 
\definecolor{tumblack}{rgb}{0, 0, 0}
\definecolor{tumwhite}{rgb}{1, 1, 1}
\definecolor{tumbluedark}{RGB}{0,82,147} 
\definecolor{tumbluelight}{RGB}{152,198,234} 
\definecolor{tumbluemedium}{RGB}{100,160,200} 
\definecolor{tumgray}{gray}{0.6} 
\definecolor{tumgray1}{gray}{0.2} 
\definecolor{tumgray2}{gray}{0.4} 
\definecolor{tumgray4}{gray}{0.6} 

\definecolor{tumivory}{RGB}{218,215,203} 
\definecolor{tumgreen}{RGB}{162,173,0} 
\definecolor{tumorange}{RGB}{227,114,34} 
\colorlet{tumgraylight}{tumgray!25!white}
\colorlet{tumgraydark}{tumgray!50!black}
\definecolor{tumdiagramaubergine}{RGB}{105,8,90}
\definecolor{tumdiagramnavyblue}{RGB}{15,27,95}
\definecolor{tumdiagramturquoise}{RGB}{0,119,138}
\definecolor{tumdiagramgreen}{RGB}{0,124,48}
\definecolor{tumdiagramlimegreen}{RGB}{103,154,29}
\definecolor{tumdiagramyellow}{RGB}{255,220,0}
\definecolor{tumdiagramsand}{RGB}{249,186,0}
\definecolor{tumdiagramredorange}{RGB}{214,76,19}
\definecolor{tumdiagramred}{RGB}{196,72,27}
\definecolor{tumdiagramdarkred}{RGB}{156,13,22}
\definecolor{compHR}{RGB}{140, 0, 0}
\definecolor{compMR}{RGB}{199,14, 21}
\definecolor{compLR}{RGB}{243,19, 29}
\definecolor{openHR}{RGB}{183,78, 18}
\definecolor{openMR}{RGB}{244,104,29}
\definecolor{openLR}{RGB}{247,154,90}
\definecolor{light}{RGB}{255, 255,0}
\definecolor{largeLow}{RGB}{192, 192, 192}
\definecolor{sparse}{RGB}{255, 204, 153}
\definecolor{industr}{RGB}{77, 77, 77}
\definecolor{denseTree}{RGB}{2, 102, 5}
\definecolor{scatTree}{RGB}{21, 255, 21}
\definecolor{bush}{RGB}{96, 135, 38}
\definecolor{lowPlant}{RGB}{185, 220, 126}
\definecolor{paved}{RGB}{0, 0, 0}
\definecolor{soil}{RGB}{ 255, 255, 204 }
\definecolor{water}{RGB}{110, 107, 254}

\definecolor{b3}{RGB}{224, 243, 219}
\definecolor{gugs}{RGB}{0, 0, 255}
\definecolor{gu}{RGB}{0, 255, 0}
\definecolor{gsp}{RGB}{255, 0, 0}
\definecolor{gh}{RGB}{119,	172,	48}
\definecolor{gup}{RGB}{77,	190,	238}
\definecolor{p}{RGB}{156, 80, 242}

\definecolor{parula61}{RGB}{  53 ,   42  , 135}
\definecolor{parula62}{RGB}{  13,   117 ,  220}
\definecolor{parula63}{RGB}{  6 ,  167 ,  198}
\definecolor{parula64}{RGB}{  101 ,  190 ,  134}
\definecolor{parula65}{RGB}{ 217 , 186 ,   86}
\definecolor{parula66}{RGB}{ 251,	150,	14}
\makeatletter
\newenvironment{customlegend}[1][]{%
    \begingroup
    \let\addlegendimage=\pgfplots@addlegendimage
    \let\addlegendentry=\pgfplots@addlegendentry
    \pgfplots@init@cleared@structures
    \pgfplotsset{#1}%
}{%
    \pgfplots@createlegend
    \endgroup
}%
\makeatother
\usepackage[%
  xindy,
  acronym,
]{glossaries}
\glsdisablehyper
\newacronym{mtl}{MTL}{multi-task learning}
\newacronym{hse}{HSE}{Human Settlement Extent}
\newacronym{lcz}{LCZ}{Local Climate Zone}
\newacronym{lc}{LC}{land cover}
\newacronym{hbase}{HBASE}{Global Human Built-up and Settlement Extent}
\newacronym{guf}{GUF}{Global Urban Footprint}
\newacronym{ghs}{GHS}{Global Human Settlement}
\newacronym{srtm}{SRTM}{Shuttle Radar Topography Mission}
\newacronym{uhi}{UHI}{Urban Heat Island}
\newacronym{cnn}{CNN}{Convolutional Neural Network}
\newacronym{fcn}{FCN}{Fully Convolutional Network}
\newacronym{cbam}{CBAM}{Convolutional Block Attention Module}
\newacronym{ndsv}{NDSV}{Normalized Difference Spectral Vector}
\newacronym{glcm}{GLCM}{gray-level co-occurrence matrix}
\newacronym{wudapt}{WUDAPT}{World Urban Database and Access Portal Tools project}
\newacronym{gsd}{GSD}{ground sampling distance}
\newacronym{gee}{GEE}{Google Earth Engine}
\newacronym{mlgcp}{MLGCP}{manually labeled grid-based checking point}
\newacronym{aspp}{ASPP}{Atrous Spatial Pyramid Pooling}
\newacronym{utm}{UTM}{Universal Transverse Mercator}
\newacronym{roi}{ROI}{Region of Interest}
\newacronym{dl}{DL}{deep learning}
\newacronym{p2f}{P2F}{predictions as features}
\newacronym{mae}{MAE}{mean absolute error}

\usepackage[capitalise]{cleveref}

\newcommand{\loss}{\ensuremath{\mathcal{L}}}

\newcommand{\ie}{{i.e., }}
\newcommand{\eg}{{e.g., }}

\usepackage{cleveref}

\begin{document}
%
%
%
%
\title{
  Multi-task Learning for Human Settlement Extent Regression and Local Climate Zone Classification
    \IEEEcompsocitemizethanks{
    \IEEEcompsocthanksitem \IEEEauthorrefmark{1}Authors share equal contribution.
    \IEEEcompsocthanksitem This work was jointly supported by the China Scholarship Concil, by the European Research Council (ERC) under the European Union's Horizon 2020 research and innovation programme (grant agreement No. [ERC-2016-StG-714087], Acronym: \textit{So2Sat}), by the Helmholtz Association through the Framework of Helmholtz Artificial Intelligence - ``Munich Unit @Aeronautics, Space and Transport (MASTr)'' and Helmholtz Excellent Professorship ``Data Science in Earth Observation - Big Data Fusion for Urban Research'' and by the German Federal Ministry of Education and Research (BMBF) in the framework of the international future AI lab "AI4EO ". The work of L. Liebel was supported by the Federal Ministry of Transport and Digital Infrastructure (BMVI) under reference: 16AVF2019A. \emph{(Correspondence: Xiao Xiang Zhu)}.
    \IEEEcompsocthanksitem C. Qiu, L. H. Hughes and M. Schmitt was with Signal Processing in Earth Observation (SiPEO), Technical University of Munich (TUM), Germany (E-mails: chunping.qiu@tum.de; lloyd.hughes@tum.de; m.schmitt@dlr.de). L. Liebel and M. Körner are with Remote Sensing Technology (LMF), TUM, Germany (E-mails: lukas.liebel@tum.de; marco.koerner@tum.de). X. X. Zhu is with the Remote Sensing Technology Institute (IMF), German Aerospace Center (DLR) and with Signal Processing in Earth Observation (SiPEO), Technical University of Munich (TUM), Germany. (E-mail: xiaoxiang.zhu@dlr.de)} 
}

\author{%
  \IEEEauthorblockN{%
    Chunping Qiu\IEEEauthorrefmark{1},
    Lukas Liebel\IEEEauthorrefmark{1}, \IEEEmembership{Student Member, IEEE},
    Lloyd H. Hughes, \IEEEmembership{Student Member, IEEE},
    Michael Schmitt, \IEEEmembership{Senior Member, IEEE},
    Marco K\"orner,
    and Xiao Xiang Zhu, \IEEEmembership{Senior Member, IEEE}%
  }
}

\maketitle

\begin{abstract}
\textcolor{blue}{\textit{This work has been accepted by IEEE GRSL for publication.}} \Gls{hse} and \gls{lcz} maps are both essential sources, \eg for sustainable urban development and \gls{uhi} studies.
Remote sensing (RS)- and \gls{dl}-based classification approaches play a significant role by providing the potential for global mapping. 
However, most of the efforts only focus on one of the two schemes, usually on a specific scale.
\textcolor{black}{This leads to} unnecessary redundancies, since the learned features could be leveraged for both of these related tasks.
In this letter, the concept of \gls{mtl} is introduced to \gls{hse} regression and \gls{lcz} classification for the first time.
We propose \textcolor{black}{a} \gls{mtl} framework and develop an end-to-end \gls{cnn}, which consists of a backbone network for shared feature learning, attention modules for task-specific feature learning, and a weighting \textcolor{black}{strategy} for balancing the two tasks.
We additionally propose to exploit \gls{hse} predictions as a prior for \gls{lcz} classification to enhance the accuracy.
The \gls{mtl} approach was extensively tested with Sentinel-2 data of 13 cities across the world.
The results demonstrate that the framework is able to provide a competitive solution for both tasks. 
\end{abstract}


\section{Introduction}
\label{sec:intro}
\IEEEPARstart{T}{}wo important tasks in urban mapping are distinguishing urban areas from non-urban background and characterizing intra-urban heterogeneity.
\Gls{hse} and \glspl{lcz} are two schemes for the respective representations.
\Gls{hse} density, indicating the \textcolor{black}{portion} of buildings, roads, and other man-made structures in an area \textcolor{black}{(e.g., a pixel)}, depicts the human footprint on Earth. By contrast,
the \gls{lcz} scheme, originally proposed for \gls{uhi} studies, includes 17 classes for detailed \gls{lc} classification \citep{stewart2012local}, \textcolor{black}{and is universally applicable}. Up-to-date, detailed, and accurate worldwide information on \gls{hse} and \gls{lcz} \textcolor{black}{can} provide support for evidence-based decision making for various applications \textcolor{black}{\eg global climate science} \citep{corbane2017big, esch2013urban}.




\gls{dl}-based approaches for \gls{lc} classification have been attracting much attention due to their proven predictive power, and end-to-end learning abilities of complex feature representations.
\Glspl{cnn}, in particular \glspl{fcn}, have been successfully applied to many RS-based image classification and segmentation tasks \citep{zhang2019joint}.
In this study, we draw upon these successes by applying \gls{dl}-based semantic segmentation methods to \gls{lc} mapping.
However, \gls{lc} mapping, particularly \gls{lcz} classification over a large scale remains challenging due to the large intra-class variability of spectral signatures, which stems from variations in physical and cultural environmental characteristics across the world \textcolor{black}{and a limited amount of reference data} \citep{MatthiasDemuzere.2019}.

This \textcolor{black}{challenge} and the lack of fully automated approaches to it motivate our investigations into \gls{dl} models with a higher generalization ability for \gls{lc} mapping.
\textcolor{black}{As already suggested by the definition of the two schemes}, prior work also indicated that \gls{hse} and \glspl{lcz} have \textcolor{black}{close correspondence  for different study areas} \citep{stewart2012local, bechtel2016towards}.
To
exploit the complementary nature of the \gls{hse} regression and \gls{lcz} classification tasks, we propose a \gls{mtl} framework to jointly predict \gls{hse} and \glspl{lcz}, considering that 
\Gls{mtl} has been shown to be a powerful technique for improving model generalization by leveraging domain knowledge of related complementary tasks \citep{ruder2017overview, Liebel20}.
In this work, we present a feature-based \gls{mtl} system that mainly consists of a shared backbone network to capture a common representation for both \gls{hse} regression and \gls{lcz} classification, and soft-attention modules to adaptively select task-specific features.
By jointly learning both tasks, we aim to boost the prediction performance over that achieved by a single, task-specific network.

\section{Methodology}
\label{sec:methodology}

\subsection{A \textcolor{black}{General} \gls{mtl} Framework for \gls{lc} Mapping on Different Scales}
\begin{figure}

  \vspace{0.5cm}

  \centering
  \begin{tikzpicture}
    \tiny

    \node (input) [inner sep=0] at (0, 0) {\includegraphics[width=1.25cm, height=1.25cm]{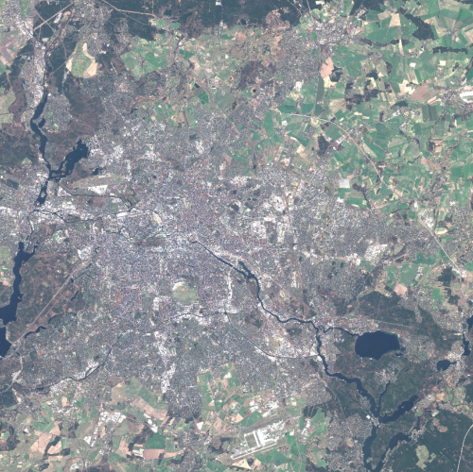}};
    \node (input_caption) [above=0cm of input] {Input Image};

    \node (lbl1) [inner sep=0] at (7.6cm, .8cm) {\includegraphics[width=1.25cm, height=1.25cm]{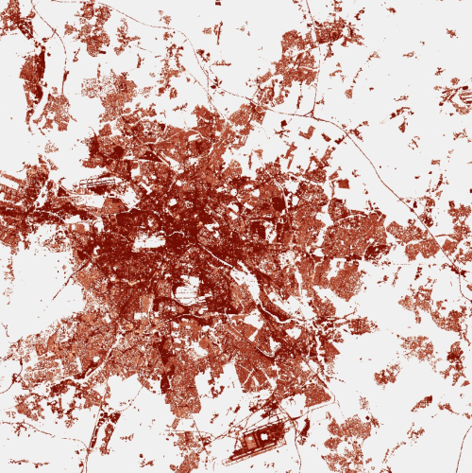}};
    \node (lbl2) [inner sep=0] at (7.6cm, -.8cm) {\includegraphics[width=1.25cm, height=1.25cm]{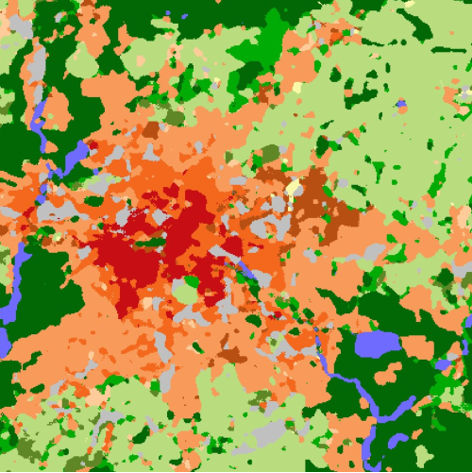}};
    \node (lbl_caption) [above=0cm of lbl1] {Labels};

    \node (pred1) [inner sep=0, left=1cm of lbl1] {\includegraphics[width=1.25cm, height=1.25cm]{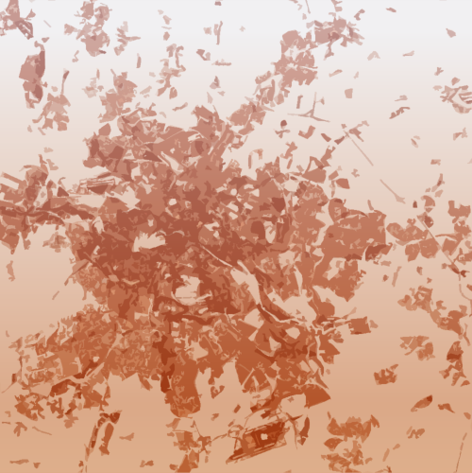}};
    \node (pred2) [inner sep=0, left=1cm of lbl2] {\includegraphics[width=1.25cm, height=1.25cm]{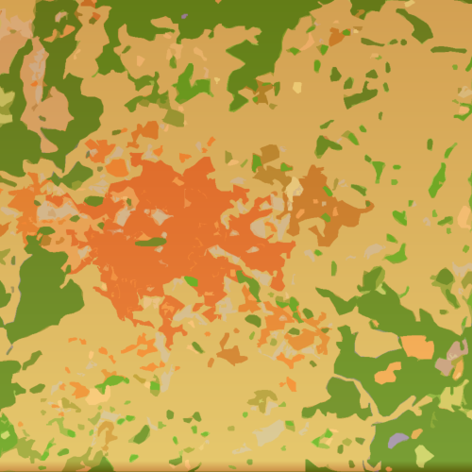}};
    \node (pred_caption) [above=0cm of pred1] {Predictions};

    \node (l1) [fill=tumgraylight, minimum width=.65cm, minimum height=.4cm] at ($(lbl1)!0.5!(pred1)$) {$\loss_{\text{HSE}}$};
    \node (l2) [fill=tumgraylight, minimum width=.65cm, minimum height=.4cm] at ($(lbl2)!0.5!(pred2)$) {$\loss_{\text{LCZ}}$};

    \node (lmt) [fill=tumgraylight, minimum width=.65cm, minimum height=.4cm, below=0.5cm of l2] {$\loss_{\text{MT}}$};

    \node (backbone) [fill=tumgraylight, right=.3cm of input] {\parbox{1.5cm}{\centering Backbone network \\ for shared \\ feature learning}};

    \node (decoder1) [fill=tumorange, left=.3cm of pred1, minimum height=.4cm] {Decoder};
    \node (decoder2) [fill=tumgreen, left=.3cm of pred2, minimum height=.4cm] {Decoder};

    \node (branch1) [fill=tumorange, left=.3cm of decoder1, minimum width=1.7cm, minimum height=.4cm] {HSE-specific module};
    \node (branch2) [fill=tumgreen, left=.3cm of decoder2, minimum width=1.7cm, minimum height=.4cm] {LCZ-specific module};

    \draw [thick, tumgray, ->] (input) -- (backbone);

    \draw [thick, tumgray, ->] (backbone.east) -| (branch1.320);
    \draw [thick, tumgray, ->] (backbone.east) -| (branch2.40);
    \draw [thick, ->, tumorange, densely dotted] (backbone.140) |- (branch1.west);
    \draw [thick, ->, tumgreen, densely dotted] (backbone.220) |- (branch2.west);

    \draw [thick, tumorange] (branch1) -- (decoder1);
    \draw [thick, ->, tumorange] (decoder1) -- (pred1);
    \draw [thick, tumgreen] (branch2) -- (decoder2);
    \draw [thick, ->, tumgreen] (decoder2) -- (pred2);

    \draw [thick, ->, tumgray] (pred1) -- (l1);
    \draw [thick, ->, tumgray] (lbl1) -- (l1);
    \draw [thick, ->, tumgray] (pred2) -- (l2);
    \draw [thick, ->, tumgray] (lbl2) -- (l2);
    \draw [thick, tumgray] (l1) -- (l2);
    \draw [thick, ->, tumgray] (l2) -- (lmt);
    \draw [thick, ->, densely dashed, draw=tumgray] (lmt) -- node [midway, above, xshift=-0.5cm] {Backpropagation for updates} ($(lmt) - (5.5cm, 0)$);

    \draw [thick, densely dotted, tumgreen] (lbl1) |- ($(pred1)!0.5!(pred2)$);
    \draw [thick, densely dotted, tumgreen] (pred1) -- ($(pred1)!0.5!(pred2)$);
    \draw [thick, ->, densely dotted, tumgreen] ($(pred1)!0.5!(pred2)$) -| (decoder2);

  \end{tikzpicture}

	\caption{\footnotesize \textcolor{black}{A general} MTL framework for HSE density regression and LCZ classification, consisting of a backbone network, task-specific network branches, and decoder modules. The inputs for network training are images and corresponding reference labels for each task.}
	\label{fig:framework_mt}
\end{figure}
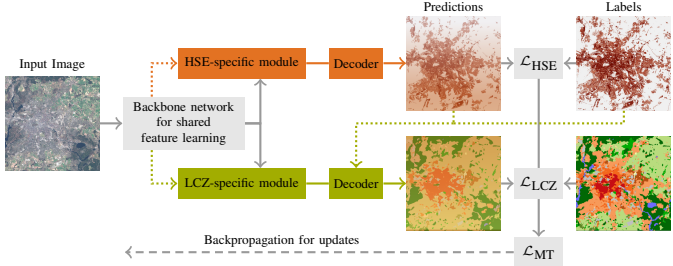

\textcolor{black}{In the definition of \gls{lcz}, one important factor to consider is the ratio of impervious surface and buildings, which indicates that the \gls{hse} density affects the categorization of an area in the \gls{lcz} scheme.} \textcolor{black}{Additionally,} \gls{hse} regression and \gls{lcz} classification prioritize spatial and semantic resolution in a complementary manner.
The \gls{lcz} scheme places emphasis on \textcolor{black}{the characterization of urban morphology} with 17 classes \textcolor{black}{(for a relatively large neighboring area, e.g., \SI{100}{\meter})}, while the \gls{hse} density contains fewer semantic details
but features a higher spatial resolution \textcolor{black}{(e.g., \SI{20}{\meter})}.
The complementary relationship, \textcolor{black}{when appropriately exploited}, is expected to support the learning process of each task and improve\textcolor{black}{s} the efficient usage of available training data.
This will potentially lead to faster convergence during training, higher accuracy for both tasks compared to single-task models, and reduced production time as both tasks can occur in parallel. To this end, a generalized \gls{mtl} framework, illustrated in \cref{fig:framework_mt}, is proposed, which consists of the following four primary components.
\begin{itemize}
  \item A backbone network for shared, general feature learning at the early stage.
  It can be implemented as state-of-the-art \gls{cnn} architectures, such as \textcolor{black}{residual neural networks}.
  Compared to natural images, \eg photographs typically used in computer vision experiments, the resolution of Sentinel-2 imagery is low.
  Hence, pooling \textcolor{black}{should be less used} within the backbone network to retain the spatial resolution \textcolor{black}{and the information of the pixel values}.
  \item Task-specific network branches for dedicated feature learning to adaptively select features from the common representations and further learn specific and high-level features for the individual tasks.
  They can be implemented as convolutional operations, attention modules or ``Squeeze-and-Excitation'' blocks \citep{hu2018squeeze}.
  \item Decoder modules to recover the required resolution of predictions. They can be implemented as transposed convolutions or upsampling layers.
  To retain clear boundaries, low level features might need to be considered, \eg via skip connections.
  \item Modules to exploit task relation. The reference of one task can be employed as prior information to guide the prediction of the other task. In this way, consistent and accurate predictions are encouraged in the MTL framework.
\end{itemize}

\subsection{Implementation of the \gls{mtl} Framework}

\begin{figure}
	\centering
	\includegraphics[width=0.49\textwidth]{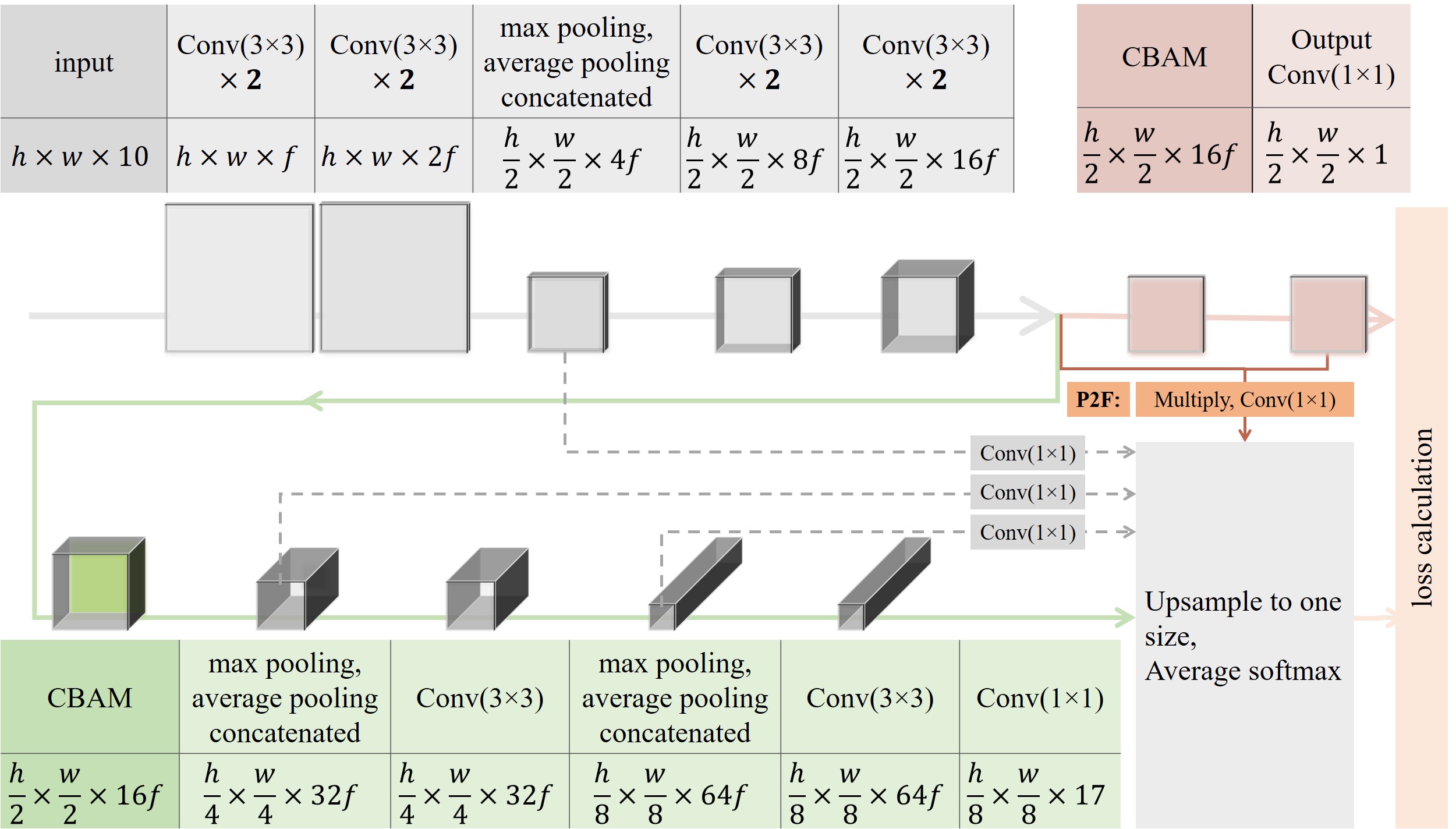}
	\caption{\footnotesize Illustration of the implemented MTL CNN architecture for HSE regression and LCZ classification. \textcolor{black}{The backbone network consists of two convolutional blocks, one pooling block, and two more convolutional blocks.} The two task-specific network branches are indicated by two different colors. The \textcolor{black}{description of each layer and the} size of inputs, feature maps, and outputs are listed \textcolor{black}{along with the operations.
$h$ and $w$ are height and width of the input patch, and $f$ is the number of feature maps from the first layer}.
	}
	\label{fig:imp_mt}
\end{figure}

One implementation of the proposed \gls{mtl} framework is illustrated in \cref{fig:imp_mt}, which consists of a backbone network \textcolor{black}{(including convolutional layers and pooling layers)} followed by two branches for \gls{hse} regression and \gls{lcz} classification.
Each of the two branches begins with an attention module, implemented as \glspl{cbam}, for an adaptive selection and learning of task-specific representations \citep{woo2018cbam}.
All $3\times3$ convolutional layers make use of separable convolution operations for the sake of efficiency.
A kernel size of $2\times2$ with a stride of 2 was used for the pooling layers, decreasing the size of feature maps by half.
Maximum and average pooling layers were used together for abstracting learned features within the architecture to preserve sufficient features.
The final output of both tasks was decided based on the desired \gls{gsd} of the respective products. 
To avoid overfitting during training, one and two drop-out layers with a dropout rate of 0.1 were utilized in the \gls{hse} and \gls{lcz} branches (omitted in the illustration).

For the \gls{hse} regression task, the last layer was activated with a sigmoid function, and 
the \gls{mae} was used as a loss function to consider potential noise in the reference data.
Considering that there are more samples \textcolor{black}{with no or few human settlements} in our dataset, a relatively high weight was assigned to samples with dense human settlements \textcolor{black}{to deal with the imbalance problem}.
The sample weight was decided based on: $e^{y_{\text{HSE}}}$, where ${y_{\text{HSE}}} \in [0,1]$ is the reference label for \gls{hse} density.

For the \gls{lcz} classification task, predicted softmax probabilities from intermediate lower-level, features were also used for the loss calculation, in addition to the final prediction.
Specifically, for each input patch, three $1\times1$ convolutional and softmax layers were used to independently predict three results using intermediate features (\textcolor{black}{indicated by dashed gray lines in \cref{fig:imp_mt}}).
This is to fully exploit the features at different levels for the elaborate \gls{lcz} scheme, which requires a diversity of representations to distinguish the 17 distinct \gls{lcz} classes.
Together with the result produced by the final layers (\textcolor{black}{represented by a solid green line in \cref{fig:imp_mt}}), these four results were upsampled to the same size as the prepared reference label patches.
All four patches of predicted softmax probabilities were averaged into one final patch for the loss calculation of the \gls{lcz} branch using a softmax cross entropy loss.

The final MTL loss, which was used to train the \gls{mtl} network, was a combination of the two single-task losses from the \gls{hse} regression task and the \gls{lcz} classification tasks.


\subsection{Predictions as Features (P2F)}

\Gls{hse} density and \glspl{lcz} do not merely share similar properties, which motivated the implicit exploitation via a \gls{mtl} framework\textcolor{black}{.} The \gls{hse} density is, furthermore, explicitly affecting the categorization of a subset of \glspl{lcz}.
For instance, an area with a high \gls{hse} density \textcolor{black}{cannot belong to a \gls{lcz} class that corresponds to natural areas}, e.g., dense trees.
Following this principle, a module called \gls{p2f} was designed to exploit this prior knowledge within our framework, the implementation of which was inspired by \citet{kohl2018probabilistic}.
The main idea is to use the \gls{hse} predictions as a prior for the classification of \glspl{lcz}, as illustrated in \cref{fig:framework_mt,fig:imp_mt}.
\textcolor{black}{Specifically, the \gls{hse} reference was multiplied with the intermediate features, resulting in processed feature maps.
These feature maps were used to get an additional prediction of \gls{lcz} classes, providing an additional output of the \gls{lcz} branch.}

The \gls{p2f} module was used in a different manner during training and test time.
Since reference labels for \gls{hse} were available at training-time, they can be utilized as the prior for the prediction of \glspl{lcz}.
At test-time, the prior was predictions of the \gls{hse} regression branch in the multi-task network, as indicated in \cref{fig:framework_mt}.
Thus, the system still solely relies on a single image as its input at test-time, while making use of the available ground-truth data during training. The quality of \gls{hse} predictions is expected to be useful for the \gls{p2f} concept as satisfying \gls{hse} mapping results have been achieved using deep \glspl{cnn} \citep{qiuFCN}.
\subsection{Dynamically Balancing Task Weights}
One of the challenges in \gls{mtl} is to balance the involved tasks, which is most commonly implemented by weighting each individual loss in the multi-task loss, subject to optimization.
Simply summing the contributing losses, assuming equal task weights of 1, is not sufficient in most cases, as this not only implies equal importance of the tasks but also that single-task losses produce values in the same order of magnitude.
Manually tuning such weights as hyper-parameters, however, is tedious.
Hence, approaches to automatically and dynamically determine them are desired and have been investigated in the literature \citep{kendall2018multi, Guo18}.

We implemented an approach to weighting the \gls{hse} regression and \gls{lcz} classification tasks based on homoscedastic uncertainty, introduced by \citep{kendall2018multi}.
Specifically, we optimize a multi-task loss function
\begin{align}
    \loss_\text{MT}(\matr W_\text{HSE/LCZ}, \sigma_\text{HSE/LCZ}) = &\frac{1}{2\sigma_\text{HSE}^2} \loss_\text{HSE}(\matr W_\text{HSE}) \nonumber \\
    &+ \frac{1}{\sigma_\text{LCZ}^2} \loss_\text{LCZ}(\matr W_\text{LCZ}) \nonumber \\
    &+ \log{\sigma_\text{HSE}} + \log{\sigma_\text{LCZ}}
\end{align}
where $\loss_{\text{HSE/LCZ}}$ is the \gls{mae} and cross entropy loss for \gls{hse} regression and \gls{lcz} classification task, respectively, $\matr W_\text{HSE/LCZ}$ are the trainable parameters from the network layers of respective branch, and $\sigma_\text{HSE/LCZ}$ are weighting parameters that control the contribution of the individual tasks. 
The variables for the input and reference data, $\matr X$, $\matr Y$, are omitted in the above equation.
The regularization terms $\log{\sigma_\text{HSE/LCZ}}$ prevent trivial solutions for $\sigma_\text{HSE/LCZ} \to \pm\infty$.
We optimized the weighting terms along with the network parameters as $s_\text{HSE/LCZ} := \log{\sigma_\text{HSE/LCZ}^2}$ due to numerical stability \citep{kendall2018multi}. 

\section{Experimental Evaluation}
\label{sec:exp}

\subsection{Dataset for Training}
A dataset was prepared to assess the potential of the proposed \gls{mtl} framework, including Sentinel-2 image patches \textcolor{black}{from three seasons (spring, summer, and autumn 2017)} and annotations for both tasks, \ie \gls{hse} density percentage and \gls{lcz} labels.

The \textcolor{black}{imagery} was processed in accordance with previous work \citep{qiuFCN}
using the five European cities, namely Berlin, Lisbon, Madrid, Milan, and Paris as study areas. Annotations for the \gls{hse} density regression are from \textcolor{black}{``High Resolution Layer Imperviousness 2015,''} using continuous values indicating the percentage of each pixel covered by \gls{hse}, \textcolor{black}{with a \gls{gsd} of \SI{20}{\meter}} \citep{CopernicusHRLI, HSEreference}.
Additionally, pixel-wise \gls{lcz} labels were prepared for each sample, using the reference from the WUDPAT project \citep{wudapLCZ}, {with an original \gls{gsd} of \SI{100}{\meter}}. It was upsampled to \SI{10}{\meter} to match the co-registered image patches during the training period.
At test time, \gls{lcz} predictions with a \gls{gsd} of \SI{100}{\meter} were produced to be consistent with state-of-the-art \gls{lcz}-related studies.
\textcolor{black}{The number of spatially disjoined training and validation patches was 75116 and 24706, respectively, with a size of $128\times128$ px.}


\subsection{Test Data and Accuracy Assessment}

For assessment of the \gls{hse} regression task, two test datasets were used.
The first one covers three test scenes in Europe, namely Amsterdam, London, and Munich and was prepared in the same way as the training data. Hence, the \gls{hse} density data is available for this test dataset.
The \gls{mae} is used as a metric when tested on this dataset.
To test the \gls{hse} regression performance on a large scale, a qualitative evaluation in comparison to HR images was carried out.
Furthermore, the predicted results were aggregated into binary labels and tested against manually annotated ground checking points, as introduced by \citet{qiuFCN}, which are uniformly distributed over ten scenes across the world.
Metrics for evaluation include Kappa, average accuracy (AA), recall, and F-Score of \gls{hse}.
This evaluation procedure, with two test datasets, can provide an estimation of the generalization ability of the proposed \gls{mtl} approach, which was trained on data from Europe exclusively.

\begin{figure}
	\centering

\pgfplotstableread{
label X1   num1  num2 num3  num4 num5  num6 num7  num8 num9  num10 num11  num12 num13 num14 num15  num16 num17
{New York}	64	526	1007	2822	144	691	3288	0	1413	3595	811	1304	332	0	332	244	215	3595
{Santiago}	36	148	197	3464	2	17	3440	0	2268	3038	144	2519	656	0	420	26	17	3037
{Jakarta}	57	123	362	2103	235	77	1132	0	788	553	505	1834	1144	17	2320	27	22	2335
{Sydney}	43	137	1639	717	285	150	638	0	1692	312	319	2	333	786	1692	216	1686	159
{Tehran}	50	114	15	2662	23	22	244	0	2316	653	265	843	100	0	1018	12	30	1330
{San Francisco}	29	359	62	1163	60	96	371	0	638	2013	40	50	170	1575	1858	161	656	0
{Rio}	22	121	132	1385	2	32	327	0	1169	164	229	713	221	7	60	176	2	1475
{Nairobi}	8	49	221	481	0	0	1155	0	951	450	72	609	10	1408	0	0	23	3
{Rome}	15	0	1794	0	0	1342	486	0	437	0	51	87	104	0	594	0	0	452
{Beijing}	1	133	391	616	596	468	410	0	613	42	29	639	119	0	596	14	5	639
}\data
\pgfplotsset{
    every tick/.style={very thin,gray},
    every tick label/.style={font={\scriptsize}},
    every axis label/.style={font={\small}},
    every axis/.append style={legend style={font=\tiny,line width=1pt,mark size=10pt}},
    xlabel style={yshift=0.2cm},
    }
\begin{tikzpicture}
    \scriptsize
        \begin{axis}
          [
            width=1\linewidth,
            height=0.45\linewidth,
            xbar stacked,
            bar width=2pt,
            ytick=data,
            yticklabels from table={\data}{label},
            ytick style={draw=none},
            xlabel={\scriptsize Number of Samples},
            xmin = 0,
            xmax = 21000,
            xtick distance=5000,
            xtick style={draw=none},
            xmajorgrids,
          ]
            \addplot+ [color=compHR] table [x=num1,y expr=\coordindex] \data;
            \addplot+ [color=compMR] table [x=num2,y expr=\coordindex] \data;
            \addplot+ [color=compLR] table [x=num3,y expr=\coordindex] \data;
            \addplot+ [color=openHR] table [x=num4,y expr=\coordindex] \data;
            \addplot+ [color=openMR] table [x=num5,y expr=\coordindex] \data;
            \addplot+ [color=openLR] table [x=num6,y expr=\coordindex] \data;
            \addplot+ [color=largeLow] table [x=num8,y expr=\coordindex] \data;
            \addplot+ [color=sparse] table [x=num9,y expr=\coordindex] \data;
            \addplot+ [color=industr] table [x=num10,y expr=\coordindex] \data;
            \addplot+ [color=denseTree] table [x=num11,y expr=\coordindex] \data;
            \addplot+ [color=scatTree] table [x=num12, y expr=\coordindex] \data;
            \addplot+ [color=bush] table [x=num13,y expr=\coordindex] \data;
            \addplot+ [color=lowPlant] table [x=num14,y expr=\coordindex] \data;
            \addplot+ [color=paved] table [x=num15,y expr=\coordindex] \data;
            \addplot+ [color=soil] table [x=num16,y expr=\coordindex] \data;
            \addplot+ [color=water] table [x=num17,y expr=\coordindex] \data;
        \end{axis}
\end{tikzpicture}
\begin{tikzpicture}
        \pgfplotsset{
        legend style={cells={anchor=west}, draw=none,column sep=1ex, nodes={scale=0.5, transform shape}}
        }
            \begin{customlegend}[legend columns=4]
            \addlegendimage{water, only marks, mark=square*}
            \addlegendentry{Water}
            \addlegendimage{soil, only marks, mark=square*}
            \addlegendentry{Bare Soil or Sand}
            \addlegendimage{paved, only marks, mark=square*}
            \addlegendentry{Bare Rock or Paved}
            \addlegendimage{lowPlant, only marks, mark=square*}
            \addlegendentry{Low Plants}
            \addlegendimage{bush, only marks, mark=square*}
            \addlegendentry{Bush (Scrub)}
            \addlegendimage{scatTree, only marks, mark=square*}
            \addlegendentry{Scattered Trees}
            \addlegendimage{denseTree, only marks, mark=square*}
            \addlegendentry{Dense Trees}
            \addlegendimage{industr, only marks, mark=square*}
            \addlegendentry{Heavy Industry}
            \addlegendimage{sparse, only marks, mark=square*}
            \addlegendentry{Sparsely Built}

            \addlegendimage{largeLow, only marks, mark=square*}
            \addlegendentry{Large Low-rise}
            \addlegendimage{light, only marks, mark=square*}
            \addlegendentry{Lightweight Low-rise}
            \addlegendimage{openLR, only marks, mark=square*}
            \addlegendentry{Open Low-rise}
            \addlegendimage{openMR, only marks, mark=square*}
            \addlegendentry{Open Mid-rise}
            \addlegendimage{openHR, only marks, mark=square*}
            \addlegendentry{Open High-rise}
            \addlegendimage{compLR, only marks, mark=square*}
            \addlegendentry{Compact Low-rise}
            \addlegendimage{compMR, only marks, mark=square*}
            \addlegendentry{Compact Mid-rise}
            \addlegendimage{compHR, only marks, mark=square*}
            \addlegendentry{Compact High-rise}

\end{customlegend}
\end{tikzpicture}

    \caption{\footnotesize Sample number of each test scene for LCZ classification assessment.}
    \label{fig:dataTestLcz}
\end{figure}
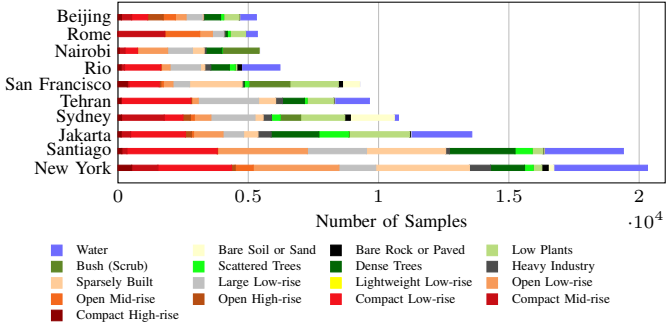

To assess the \gls{lcz} classification task, we utilized the same ten worldwide scenes as the \gls{hse} regression, with reference labels from the So2Sat LCZ42 dataset, \textcolor{black}{which was processed by polygon shrinking and class balancing after labelling }\citep{xzLcz42Data}.
The number of test samples for each scene is shown in \cref{fig:dataTestLcz}.
In addition to overall accuracy (OA), Kappa, and AA, Weighted accuracy (WA) was used \textcolor{black}{by assigning different penalties to different mistakes.}

\subsection{Hyper-Parameter Settings}

All models in this study were implemented in Keras for TensorFlow and trained from scratch. Basic hyper-parameters include a batch size of 8 and an initial learning rate of 0.002 for the Nesterov Adam optimizer.
The learning rate was decreased by 0.25 after every two epochs.
To control the training time and avoid overfitting, early stopping was implemented with the validation loss as the monitored metric with a patience of
10 epochs.

\subsection{Results for Different Settings}

\Cref{tab:compareAcc_mt} presents evaluation results for \gls{hse} regression and \gls{lcz} classification over both test sets.
In addition to single- and multi-task training, different strategies of feature exploitation and task weighting are also compared.

\begin{table*}
  \caption{\footnotesize Comparison of results from different approaches tested within and outside Europe. The values were averaged over three and ten test areas for the European and the global test set. For each metric, the top two best results were marked in bold.}
  \label{tab:compareAcc_mt}

  \scriptsize
  \sisetup{detect-weight=true, detect-inline-weight=math}

  \begin{tabularx}{\linewidth}{@{}l*{13}{S[table-auto-round, table-format=1.2]}@{}}
    \toprule

    \multicolumn{1}{@{}l}{Configuration} & \multicolumn{5}{l}{European Test Set} & \multicolumn{8}{l@{}}{Global Test Set} \\
    \cmidrule(lr){2-6} \cmidrule(l){7-14}

    & \multicolumn{1}{l}{HSE Density} & \multicolumn{4}{l}{HSE Aggregated} & \multicolumn{4}{l}{HSE Aggregated} & \multicolumn{4}{l@{}}{LCZ} \\
    \cmidrule(lr){2-2} \cmidrule(lr){3-6} \cmidrule(lr){7-10} \cmidrule(l){11-14}

    & \multicolumn{1}{X}{MAE} & \multicolumn{1}{X}{Kappa} & \multicolumn{1}{X}{AA} & \multicolumn{1}{X}{Recall} & \multicolumn{1}{X}{F} & \multicolumn{1}{X}{Kappa} & \multicolumn{1}{X}{AA} & \multicolumn{1}{X}{Recall} & \multicolumn{1}{X}{F} & \multicolumn{1}{X}{OA} & \multicolumn{1}{X}{Kappa} & \multicolumn{1}{X}{AA} & \multicolumn{1}{X}{WA} \\
    \cmidrule(r){1-1}
    \cmidrule(lr){2-2} \cmidrule(lr){3-3} \cmidrule(lr){4-4} \cmidrule(lr){5-5} \cmidrule(lr){6-6} \cmidrule(lr){7-7} \cmidrule(lr){8-8} \cmidrule(lr){9-9} \cmidrule(lr){10-10} \cmidrule(lr){11-11} \cmidrule(lr){12-12} \cmidrule(lr){13-13} \cmidrule(l){14-14}

    \addlinespace

    Single-Task HSE                       & 4.24        & 0.76        & \sbf 0.92   & 0.91        & 0.94        & 0.78        & 0.8879      & \sbf 0.9302 & 0.89        & \textemdash & \textemdash & \textemdash & \textemdash \\
    Single-Task LCZ                       & \textemdash & \textemdash & \textemdash & \textemdash & \textemdash & \textemdash & \textemdash & \textemdash & \textemdash & 0.4784      & 0.41        & 0.3413      & 0.8543      \\

    \addlinespace

    Multi-Task (1:1, CBAM)                      & 4.55        & 0.69        & 0.90        & 0.86        & 0.91        & 0.77        & 0.8855      & \sbf 0.9445 & 0.89        & \sbf 0.4951 & \sbf 0.42   & 0.3379      & 0.8773      \\
    Multi-Task (1:1, CBAM, P2F)                 & 4.39        & 0.71        & 0.90        & 0.87        & 0.92        & 0.78        & 0.8895      & 0.9221      & \sbf 0.90   & 0.4397      & 0.37        & 0.3095      & 0.8173      \\

    \addlinespace

    Multi-Task (learned weights, CBAM)          & 4.49        & 0.76        & 0.91        & 0.91        & 0.94        & \sbf 0.80   & 0.8995      & 0.9182       & \sbf 0.90  & 0.4813      & \sbf 0.42   & \sbf 0.3609 & \sbf 0.8809 \\
    Multi-Task (learned weights, P2F) & \sbf 4.14   & \sbf 0.77   & \sbf 0.92   & \sbf 0.92   & \sbf 0.95   & \sbf 0.81   & \sbf 0.9036 & 0.9238      & \sbf 0.91  & 0.4533      & 0.38        & 0.3169      & 0.8318      \\
    Multi-Task (learned weights, CBAM, P2F)     & \sbf 4.12   & \sbf 0.79   & 0.91        & \sbf 0.93   & \sbf 0.95   & \sbf 0.80   & \sbf 0.9004 & 0.9032       & \sbf 0.90  & \sbf 0.5011 & \sbf 0.44   & \sbf 0.3637 & \sbf 0.8813 \\


    \bottomrule
  \end{tabularx}
\end{table*}

\begin{figure*}
  \includegraphics[height=0.27\linewidth]{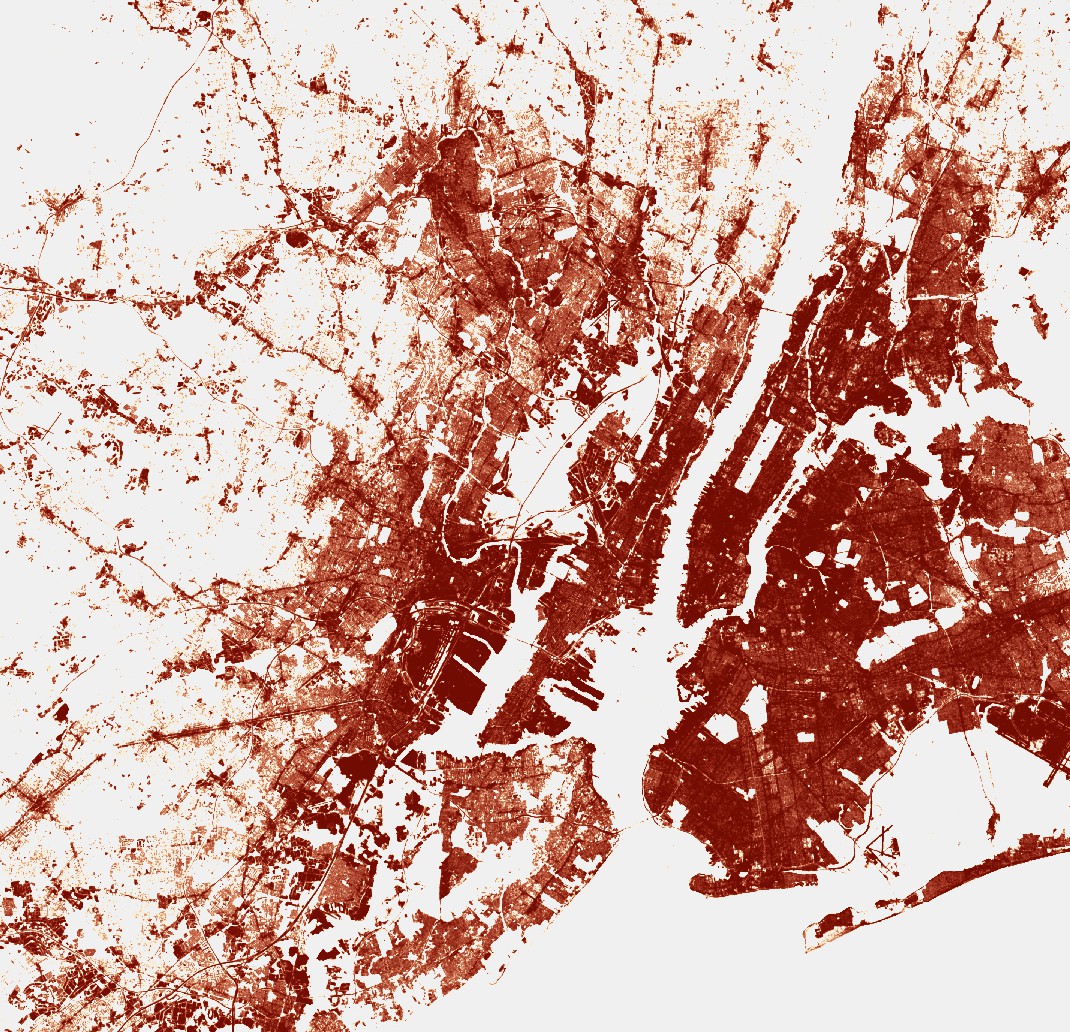}
  \includegraphics[height=0.27\linewidth]{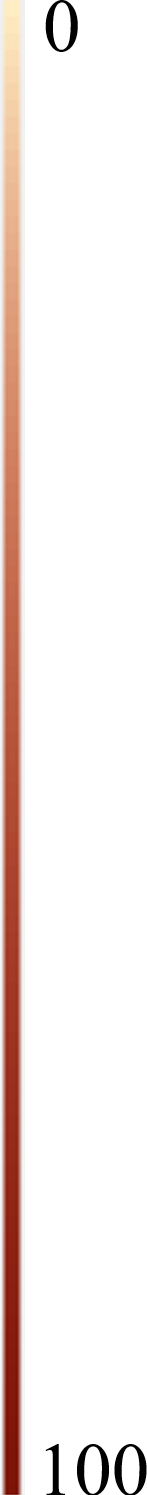}
  \hfill
  \includegraphics[height=0.27\linewidth]{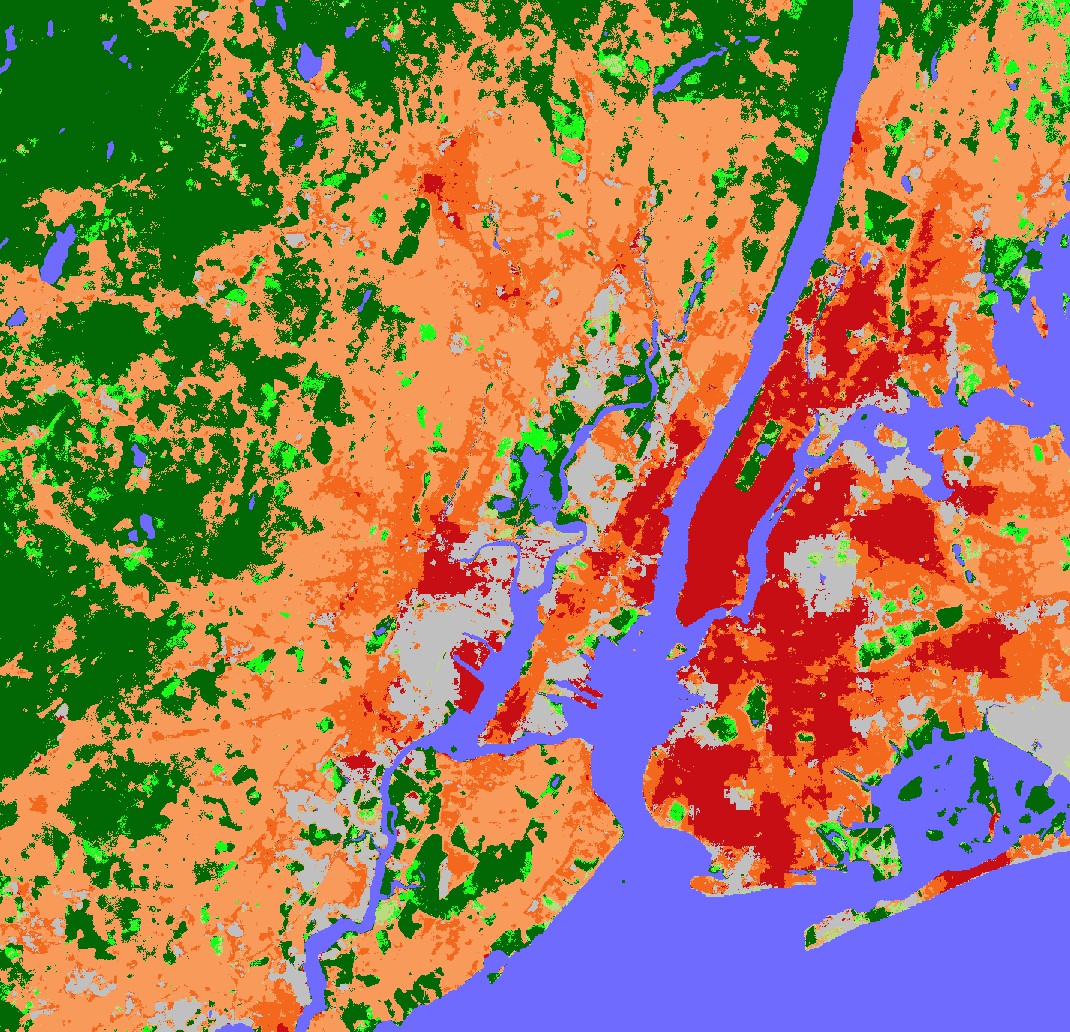}
  \hfill
  \includegraphics[height=0.27\linewidth]{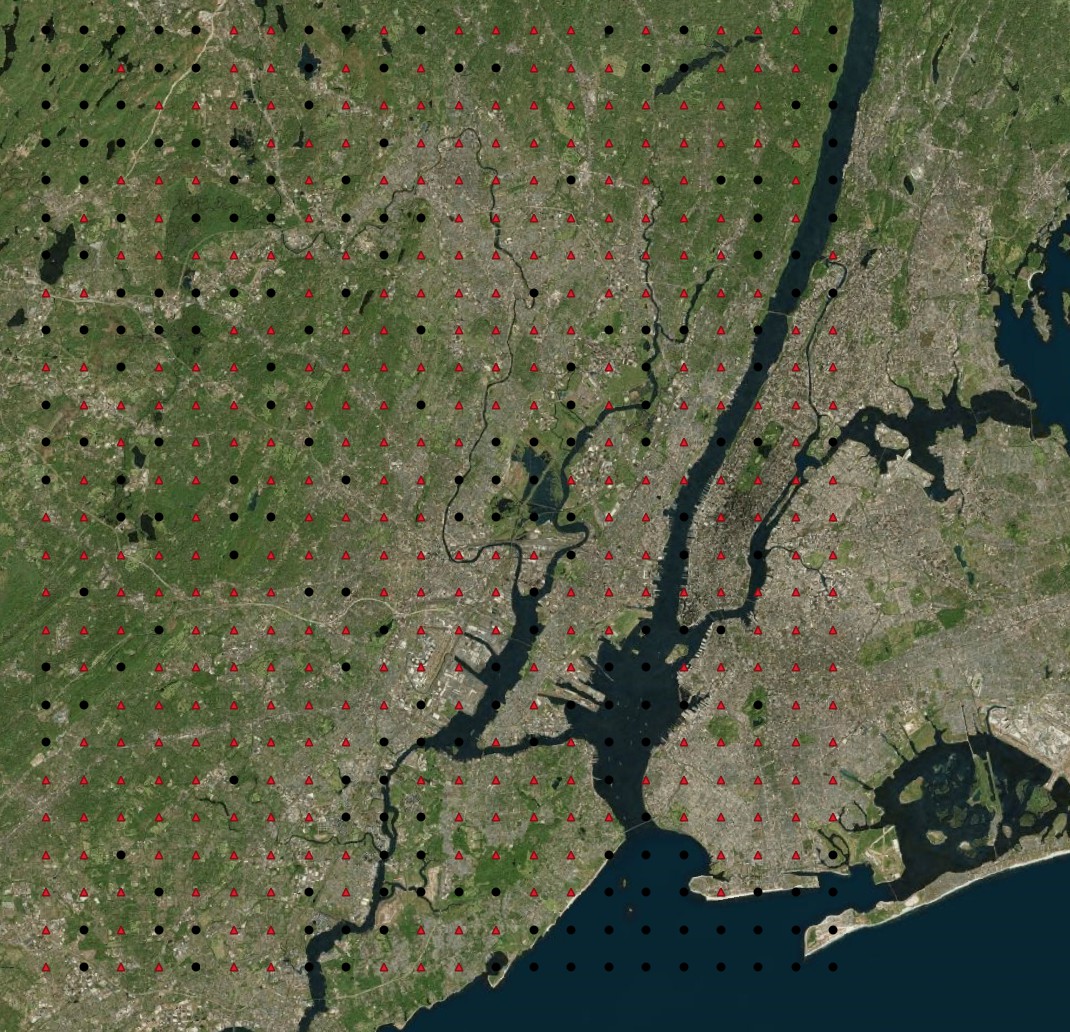}

  \caption{\footnotesize An illustration of joint prediction results in New York City (NYC), USA. From left to right are results of HSE regression and LCZ classification (\textcolor{black}{with the same legend as \cref{fig:dataTestLcz}}), and HR image overlaid with reference points (red for HSE and black for non-HSE), respectively.}
  \label{fig:ny_mt}
\end{figure*}

An illustration of the \gls{mtl} predictions is presented in \cref{fig:ny_mt} for a test scene in New York City.
Both \gls{hse} regression and \gls{lcz} classification results reflect the pattern of urban structures in the imagery, indicating a reasonable mapping result.


\section{Discussion}
\label{sec:dis}


\subsection{Superiority of the Proposed MTL Approach}

Joint prediction was able to provide benefits for both tasks, \textcolor{black}{as analyzed in the following.}
Simply weighting the two contributing single-task losses equally ($1:1$) only resulted in a slight improvement for OA and Kappa in the \gls{lcz} classification task, as shown in \cref{tab:compareAcc_mt}.
When using the dynamically learned weights, the benefit became apparent, with AA being improved from 0.89 to 0.90 and from 0.34 to 0.36 for \gls{hse} and \gls{lcz} results.
This shows that the dynamical weight learning strategy plays a positive role in the \gls{mtl} performance.

The proposed strategy of using the \gls{hse} density predictions as features for \gls{lcz} classification, \gls{p2f}, was able to additionally improve the effectiveness of the \gls{mtl} configuration, providing best results among all investigated approaches for the \gls{lcz} classification task.
This is an improvement over the basic approaches, such as single-task learning, with Kappa being improved from 0.41 to 0.44 and OA being improved from 0.478 to 0.50.
This effect can be attributed to guidance from the \gls{hse} density mapping task.
Intuitively, the candidate \gls{lcz} type of an area can be narrowed down when built-up areas were already detected in the \gls{hse} regression task.
This piece of useful information from \gls{hse} regression can be leveraged for \gls{lcz} classification, leading to a further improvement over baseline \gls{mtl} approaches.
It can also be noted from \cref{tab:compareAcc_mt} that \gls{p2f} was not able to provide improvement when the task weights were $1:1$, which indicates the importance of an appropriate weighting strategy.

Task-specific feature learning modules, \eg \glspl{cbam} utilized in the experiments, are important for gaining benefits within the \gls{mtl} framework.
This can be proved by comparing the results with and without \glspl{cbam} in Table \ref{tab:compareAcc_mt}.
All metrics for \gls{lcz} classification were worse when removing \glspl{cbam} from the \gls{mtl} architecture.
A possible reason is that \gls{hse} mapping and \gls{lcz} classification each require some different lower-level features, which cannot be satisfied in the absence of attention modules as a feature selection process.
\textcolor{black}{Another possible reason for the improved results from MTL is that multi-source reference data is used. This helps to learn more generalized features, and thus the model is less prone to overfitting, compared to single-task learning.}
\subsection{Quality of Current Mapping Results and Future Work}

The achieved \gls{hse} regression results, representing continuous \gls{hse} density, provide richer information beyond a binary delineation of urban areas, such as the \gls{guf}, the \gls{ghs} built-up grid, and our previous work \citep{qiuFCN}.
Quantitatively, the accuracy is higher using the same checking points as a test set. Specifically, the AA, recall, and F-score of our previous binary HSE mapping results are 0.90, 0.91, and 0.91, respectively. All these metrics are beyond the state-of-the-art products when tested on the same 10 distinct scenes. For instance, AA of the \gls{guf} and the \gls{ghs} built-up grid are 0.86 and 0.83.
It should be noted, however, that the temporal gaps between the checking points and these products might play a role in this comparison.

As a proof of concept, reference data from only five European scenes was collected and used to train the \gls{mtl} network.
Still, the achieved \gls{lcz} classification results are promising over ten distinct test scenes across the world, demonstrating the capability for generalization \textcolor{black}{and potential for further exploring this challenging task}.
The accuracy is reasonable when compared to state-of-the-art work using similar experimental settings, \textcolor{black}{i.e., tested on completely unseen areas} \citep{MatthiasDemuzere.2019}.
More accurate \gls{lcz} classification results can be expected if \textcolor{black}{larger} amounts of high-quality reference data \textcolor{black}{are} available in the future, as the \gls{lcz} reference might still contain errors, even after manual editing by expert annotators.
\textcolor{black}{Future work includes exploring the potential of \gls{mtl} for more test scenes and longer time series to provide support for applications such as environmental management and monitoring.}
\section{Conclusion}
\label{sec:con}
\gls{hse} density and \gls{lcz} maps are both vital for urban analysis.
Based on the intuition that both tasks are highly correlated and might provide hints for each other, we proposed to jointly predict \gls{hse} density and \gls{lcz} labels via a \gls{mtl} approach, and developed a \gls{mtl}-based \gls{cnn} \textcolor{black}{to better leverage the complimentary features extracted from the input satellite imagery}.
We validated the proposed approach with extensive experiments using Sentinel-2 data in the context of large-scale urban analysis. 
\textcolor{black}{This letter shows that related urban mapping tasks can be performed jointly for improved generalization ability of \gls{dl} models.}

\ifCLASSOPTIONcaptionsoff
  \newpage
\fi



%

\small
\bibliographystyle{IEEEtranN}
\bibliography{ref.bib}

%




\end{document}